\documentclass{article}

\usepackage[final]{neurips_2019}

\usepackage{amsthm, amsmath, amssymb}
\usepackage[utf8]{inputenc} % allow utf-8 input
\usepackage[T1]{fontenc}    % use 8-bit T1 fonts
\usepackage{setspace}
\usepackage{paralist}
\usepackage{hyperref}
\hypersetup{colorlinks = true,
            linkcolor = black,
            urlcolor  = black,
            citecolor = blue,
            anchorcolor = black}
\setcitestyle{round}

\usepackage{graphicx}
\usepackage{url}            % simple URL typesetting
\usepackage{booktabs}       % professional-quality tables
\usepackage{amsfonts, bm}   % blackboard math symbols & bold math
\usepackage{nicefrac}       % compact symbols for 1/2, etc.
\usepackage{microtype}      % microtypography

\newtheorem{conj}{Conjecture}[section]

\newcommand{\set}[1]{\{#1\}}

\title{Sparse Transfer Learning via Winning Lottery Tickets}

\author{%
  Rahul S.~Mehta\thanks{Palantir Technologies, 15 Little W. 12th Street, New York, NY, 10004. This research was supported by Princeton University, Amazon Research, and the Arora Group. This work was completed while the author was a student at Princeton University.} \\
  Palantir Technologies \\
  \texttt{rahulm@palantir.com} \\
  }

\begin{document}

\maketitle

\begin{abstract}
The recently proposed Lottery Ticket Hypothesis of \cite{Frankle18} suggests that the performance of over-parameterized deep networks is due to the random initialization seeding the network with a small fraction of favorable weights. These weights retain their dominant status throughout training -- in a very real sense, this sub-network ``won the lottery" during initialization. The authors find sub-networks via unstructured magnitude pruning with 85-95\% of parameters removed that train to the same accuracy as the original network at a similar speed, which they call {\it winning tickets}. In this paper, we extend the Lottery Ticket Hypothesis to a variety of transfer learning tasks. We show that sparse sub-networks with approximately 90-95\% of weights removed achieve (and often exceed) the accuracy of the original dense network in several realistic settings. We experimentally validate this by transferring the sparse representation found via pruning on CIFAR-10 to SmallNORB and FashionMNIST for object recognition tasks.
\end{abstract}

\section{Introduction}
Deep neural networks (DNNs) achieve state-of-the-art performance for a wide array of machine learning and pattern recognition problems. Conventional wisdom has long held that increasing the depth of a neural network improves its expressive power. This overparameterization also seems to {\it improve} the generalization ability of deep networks, which is at odds with traditional machine learning methods. Despite these benefits, increased depth also complicates optimization; deep networks encounter issues such as convergence failures, and are often prohibitively expensive in resource-constrained settings. Conversely, techniques for removing unnecessary weights from networks such as pruning \citep{Cun90optimalbrain, HanPTD15, LiKDSG16} can eliminate up to 85-90\% of parameters with no effect, but still require training the full model. 

Recently, \cite{Frankle18} proposed the {\it lottery ticket hypothesis} (LTH), which states that there exists a sparse, {\it trainable} sub-network within a dense feed-forward network that reaches the same validation accuracy as the original network when trained from scratch.

\paragraph{The Lottery Ticket Hypothesis} A randomly-initialized, dense feed-forward neural network contains a subnetwork that is initialized such that -- when trained in isolation -- it can match the test accuracy of the original network after training for at most the same number of iterations.

Formally, consider a dense feed-forward network $f(\bm{x}; \bm{\theta})$ with initial parameters $\bm{\theta}_0 \sim \mathcal{D}_\theta$. Over the course of optimization, $f$ reaches minimum validation loss $\ell$ at iteration $j$, with a test accuracy $\alpha$. Also consider the same network $f$ re-trained from $f(\bm{x};m \odot \bm{\theta}_0)$, for some $m \in \set{0,1}^{|\bm{\theta}|}$ (it is clear that as long as $||m||_0 > 0,$ $||m \odot \bm{\theta}||_0 < |\bm{\theta}|$), and suppose that it reaches its minimum validation loss $\ell'$ at iteration $j'$ with test accuracy $\alpha'$.\footnote{In this context, $\odot$ denotes the element-wise (Hadamard) product.} The Lottery Ticket Hypothesis formally states that $\exists\: m \in \set{0,1}^{|\bm{\theta}|}$ such that $j' \leq j$ ({\it commensurate training time}) and $\alpha' \geq \alpha$ ({\it commensurate accuracy}). We call the sub-network, defined by $m$, a {\it winning ticket}. All winning tickets can be found by simple unstructured magnitude pruning \citep{HanPTD15}. We note that while this procedure still requires that the dense network be fully-trained, the sparse sub-network can be re-trained from scratch to reach the same test accuracy or even {\it exceed} that of the original network in the same number of training steps. 

Subsequently, \cite{frankle19scale} found that such winning tickets can be scaled up to large visual recognition datasets such as ImageNet. However, this line of research still leaves one important question open; does the structure of winning tickets depend on the r they were trained on? \cite{Frankle18} allude to this possibility when discussing the implications of their conjecture, and hypothesize that winning tickets may be transferrable between tasks.

\paragraph{Contributions} This paper focuses on investigating whether or not the sparse representation found via unstructured magnitude pruning can be trained in isolation on a {\it distinct} target task. In particular, we show the following results:\footnote{We make our source code available at \href{ https://github.com/rahulsmehta/sparsity-experiments}{\tt https://github.com/rahulsmehta/sparsity-experiments}.}

\begin{itemize}
    \item We pose the {\it Ticket Transfer Hypothesis}, modifying the original LTH for transfer learning problems. We formalize this statement in Conjecture \ref{conj:lth-transfer}. In doing so, we resolve several inconsistencies between the original statement of the conjecture and the transfer learning setting.
    
    \item We show that sparse representations with as few as 5\% of parameters remaining found via pruning on CIFAR-10 can be transferred to SmallNORB and FashionMNIST when only the dense feed-forward layers are fine-tuned on the target task (which mirrors many realistic transfer learning scenarios).
    
    \item We repeat the above experiments, this time fine-tuning the {\it entire} network; we find similar winning tickets with up to 95\% of parameters removed that reach or exceed the performance of the original network. We also investigate the efficacy of the {\it iterative} and {\it one-shot} pruning methods introduced by \cite{HanPTD15}. 
\end{itemize}

% We also open-source the code for our experiments. In particular, we provide a PyTorch implementation for network pruning/finding winning tickets as well as layer-wise sensitivity analysis.\footnote{\href{ https://github.com/rahulsmehta/sparsity-experiments}{\tt https://github.com/rahulsmehta/sparsity-experiments}} % TODO: FIX LINK

\section{Revisiting the Lottery Ticket Hypothesis}\label{sec:lth_transfer}
The original Lottery Ticket Hypothesis is proposed in the context of a single learning task $\mathcal{T}$; the dense network is trained to completion on the task, and a sparse sub-network is found via iterative pruning. Then, the sparse sub-network is reset to the original weights and retrained in isolation on $\mathcal{T}$. 

In contrast, in the {\it transfer learning} setting, the representation learned by a network on a {\it source} task $\mathcal{T}_S$ is leveraged when fine-tuning the model on a distinct {\it target} task $\mathcal{T}_T$. For instance, if we wish to classify a target dataset with relatively few labeled examples, one approach is to transfer the weights learned from a more general task such as object detection on ImageNet, and then fine-tune the representation on the target dataset. In this setting, it is natural to ask if winning tickets found via iterative pruning on a source task $\mathcal{T}_S$ can reach commensurate accuracy when fine-tuned in isolation on a target task $\mathcal{T}_T.$

Recall that the original statement of the Lottery Ticket Hypothesis specifies that the network can be reset to its original initialization $f(\bm{x}; m \odot \bm{\theta}_0)$ after $m$ is found via pruning. However, this does not make sense in a transfer learning context; if the main motivation is to transfer a learned feature representation from a large dataset such as ImageNet to a task with few labelled examples, resetting the network to $f(\bm{x}; m \odot \bm{\theta}_0)$ eliminates any information learned from the source task; in a sense, the ``winning ticket" initialization should be equivalent to a random initialization $f(\bm{x}; m \odot \bm{\theta}')$ for some $\bm{\theta}' \sim \mathcal{D}_\theta$. Moreover, we can expect that re-initializing the network to any other early weights $\bm{\theta}_{j'}$ ($j' \ll j$) will adversely affect performance, and prevent the sub-network from reaching the original network's accuracy. This should be especially evident in the case of only fine-tuning the fully-connected layers while freezing the convolution layers (recall that $j$ is the iteration of minimum validation loss), since this is essentially equivalent to using a fixed random feature embedding throughout training.

Accordingly, we first informally state our adaption of the Lottery Ticket Hypothesis for transfer learning, which we call the {\it Ticket Transfer Hypothesis}. We conveniently ignore the matter of initialization below for our informal description of the conjecture; it will be easier to rigorously define this in Conjecture \ref{conj:lth-transfer}.

\paragraph{Ticket Transfer Hypothesis (Informal)}  Given a source and target task (i.e. classification), a randomly-initialized, dense feed-forward neural network contains a subnetwork that can be found while training on the source task such that -- when fine-tuned in isolation on a target task -- it can match the test accuracy of the original network fine-tuned on the target task in as many training iterations. These sub-networks can always be found with unstructured magnitude pruning.

The major distinction between our variant and the original formulation of the LTH is that the sparse sub-network must be determined using only the {\it source} dataset; moreover, we require that the sub-network reaches the same accuracy in the same number of training setps when fine-tuned on the target task as the original dense network.

In order to address the matter of initialization, we adapt the concept of {\it late resetting}. In this approach, instead of re-initializing $f$ to $m \odot \bm{\theta}_0,$ $f$ is reset to some $m \odot \bm{\theta}_{j'}$ ($j' \ll j$). \cite{frankle19scale} show that late resetting produces winning tickets that can overcome several stability issues and scale to large-scale tasks such as ImageNet. Moreover, in the classic transfer learning setting, the final fitted representation is always the initialization for a pre-trained network fine-tuned on a downstream task. Thus, we consider this to be the analog of late resetting in the transfer learning setting.

We make this intuition precise below; in particular, we resolve the issue of initialization by relaxing the condition that $f(\bm{x}; m \odot \bm{\theta}_0)$ must reach commensurate accuracy on the target task. Consider source and target tasks $\mathcal{T}_S$ and $\mathcal{T}_T$. Let $f(\bm{x},\bm{\theta})$ be a dense, feed-forward network initialized to $\bm{\theta} = \bm{\theta}_0 \sim \mathcal{D}_\theta$. Let $j$ denote the iteration at which $f$ reaches its minimum validation loss on $\mathcal{T}_S$. Further, let $\mathcal{P}_S = \set{\bm{\theta}_i}_{i=0}^j$ denote the set of parameters through the $j$th training iteration in the source task $\mathcal{T}_S$, and let $\bm{\theta}_S$ denote the parameters at the iteration of minimum validation loss. Moreover, suppose that $f(\bm{x}; \bm{\theta}_S)$ reaches its minimum validation loss $\ell$ and test accuracy $\alpha$ on $\mathcal{T}_T$ at iteration $k$.

\begin{conj}[Ticket Transfer Hypothesis]\label{conj:lth-transfer}
Suppose that $f(\bm{x};m \odot \bm{\theta}^*)$ reaches minimum validation loss $\ell'$ at iteration $k'$ with test accuracy $\alpha'$ on $\mathcal{T}_T$. Then there exists $m \in \set{0,1}^{|\bm{\theta}|}$, $\bm{\theta}^* \in \mathcal{P}_S$ such that $||m||_0 \ll |\bm{\theta}|, k' \le k$ (commensurate training time), and $\alpha' \geq \alpha$ (commensurate accuracy).
\end{conj}

We address the issue of initialization by enabling our sparse sub-network to be reset to {\it any} weights that were encountered during the optimization trajectory of the model. This can be viewed as a generalization of a ``winning ticket" initialization -- since we would like to transfer the representation to another task, we are satisfied if we can achieve commensurate accuracy with {\it any} initialization found during training on the source task $\mathcal{T}_S$. In practice, we consider three initializations; $\bm{\theta}_0$ (the ``winning ticket" initialization), $\bm{\theta}_S$ (``late reset"), and $\bm{\theta}' \sim \mathcal{D}_\theta$ (random).

\section{Transferring Winning Tickets for Object Recognition}
In the remainder of this paper, we investigate the Ticket Transfer Hypothesis for object recognition problems. Using unstructured magnitude pruning, we find winning tickets that retain between 5-15\% of the original parameters when transferring a sparse the representation found on CIFAR-10 to SmallNORB and FashionMNIST. In this section, we find winning tickets when the convolution layers are frozen while fine-tuning on the target task. Below, we briefly outline our algorithm for transferring winning tickets:
\begin{compactenum}
\item Randomly initialize $f(\bm{x};\bm{\theta})$ and train the network on the source task $\mathcal{T}_S$. Let $\bm{\theta}_S$ denote the parameters at the iteration of minimum validation loss.
\item Over $k$ iterations, iteratively prune the network until $p$\% of the parameters (the desired sparsity level) remain. Let $m \in \set{0,1}^{|\bm{\theta}|}$ denote the parameter mask.
\item Reset the pruned network to $f(\bm{x}; m \odot \bm{\theta}_S)$, and train $f$ on the target task $\mathcal{T}_T$
\end{compactenum}

% Our results are organized as follows: Section \ref{sec:vision_models} briefly describes the model architectures, and gives a brief overview of our datasets and our explicit transfer learning tasks. Section \ref{sec:vision_lt_freeze} shows how to find winning tickets when the convolution layers are frozen during fine-tuning, and Section \ref{sec:vision_lt_ft_all} repeats these experiments when the entire network is fine-tuned.
For our vision experiments, we investigate two widely-used convolutional networks; ResNet18 and VGG19. We also compare the performance of winning ticket and random initializations to two simple fully-connected architectures. For convolutional networks, we use the same configurations for our experiments as \cite{Frankle18}. These configurations are summarized in Table \ref{tab:vision_models}.

For convolutional networks, we eliminate 20\% of parameters from the convolution layers at each iteration using unstructured magnitude pruning. During pruning, we leave the fully-connected layers untouched. This is critical in the context of transfer learning, since in some settings we only fine-tune the fully-connected layer while leaving the pruned convolution layers frozen.

In our various experiments, we use CIFAR-10 \citep{cifar} as our source dataset, and fine-tune our models for two target datasets: FashionMNIST \citep{fmnist} and SmallNORB \citep{norb}, which we will abbreviate as FMNIST and NORB respectively. The source and target task is multi-class classification. The various attributes of each dataset are summarized in Table \ref{tab:vision_datasets}. We apply standard data-augmentation when training and fine-tuning our models: specifically, we augment inputs at training time using random horizontal flips and random 4-pixel pads and crops.

\subsection{Winning Tickets with Frozen Convolutions}\label{sec:vision_lt_freeze}
In this section, we consider fine-tuning winning tickets found for CIFAR-10 to both SmallNORB and FMNIST. We consider the setting where we ``freeze" the convolution layers while fine-tuning on the target task, and only train the fully-connected layers. 

We also consider two variants of this setting; first, we only fine-tune the single fully-connected layer in ResNet18 after training on the source task. Next, we replace the final fully-connected layer with a 2-layer feed-forward network, which is common practice for a wide variety of transfer learning tasks \citep{pan2010survey}. In our experiments, we train sparse sub-networks from 3 different initializations; (1) the {\it late-reset} initialization, which in the transfer learning context corresponds to the fitted weights $\bm{\theta}_S$ at the end of training on the source task, (2) the {\it winning ticket} initialization $\bm{\theta}_0$, which is the original initialization for training on the source task, and (3) {\it randomly re-initializing} the weights to some $\bm{\theta}' \sim \mathcal{D}_\theta$ before fine-tuning.

First, we show results for transferring winning tickets found on CIFAR-10 to SmallNORB. We initially consider the setting in which we only fine-tune the existing dense layer in ResNet18. Intuitively, this treats the pruned convolution layers as a sparse feature representation, and fine-tunes the classifier to predict the labels based on this representation. 

\begin{figure}[h]
    \centering
    \includegraphics[width=0.75\textwidth]{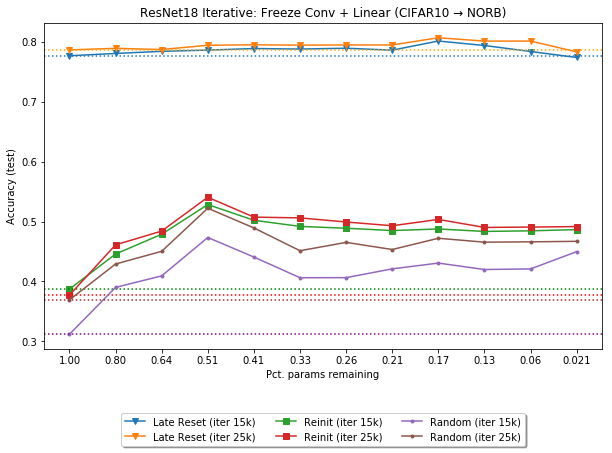}
    \caption[Winning Tickets: CIFAR-10 to NORB (Freeze + Linear)]{Winning Tickets: CIFAR-10 to NORB (ResNet18, freeze conv. + linear)}
    \label{fig:ticket_freeze_linear_norb}
\end{figure}

Figure \ref{fig:ticket_freeze_linear_norb} summarizes our results. In particular, we note that when fine-tuning from the late-reset initialization, we find winning tickets with as few as 6\% of original parameters remaining. The top test accuracy of 80.01\% is achieved when fine-tuning a winning ticket with 17\% of original parameters retained.

Both the winning ticket and random reinitialization fails to reach commensurate accuracy at all pruning levels. In both cases, though, test accuracy improves as we sparsify the convolution layers. Both the random and winning ticket initializations reach their respective maximum test accuracies with 51\% of parameters remaining. Accuracy remains relatively flat at that point, despite pruning more weights in the convolution layers. 

\begin{figure}[h]
    \centering
    \includegraphics[width=0.75\textwidth]{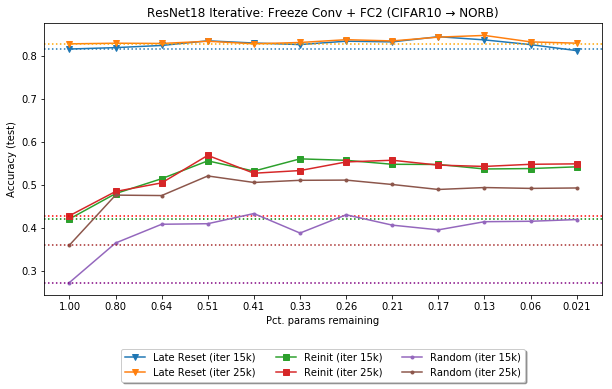}
    \caption[Winning Tickets: CIFAR-10 to NORB (Freeze + FC2)]{Winning Tickets: CIFAR-10 to NORB (ResNet18, freeze conv. + 2-layer fully-connected)}
    \label{fig:ticket_freeze_fc2_norb}
\end{figure}

Next, we repeat the experiment for ResNet18 after replacing the single fully-connected layer with a 2-layer network before fine-tuning. This increases the pruned parameter count by approximately 50,000 parameters, which is negligible when compared to the size of the convolution layers. Figure \ref{fig:ticket_freeze_fc2_norb} summarizes our results.

We demonstrate the same effect as in the linear output layer case; when fine-tuned on the target task when initialized with the fitted weights at the end of the source training task $\bm{\theta}_S$, we find winning tickets with as few as 6\% of parameters retained. Note that while this is at the same sparsity level for the original winning ticket as before, we reach a superior test accuracy of 84.41\%. Since we do not prune the dense layers, the acceleration effect described by \cite{AroraOprm18} likely accounts for the improvement in accuracy due to the overparameterization in the fully-connected layers. This is further supported by empirical evidence presented by \cite{mehtaOprm}.

\begin{figure}[h]
    \centering
    \singlespacing
    \includegraphics[width=0.75\textwidth]{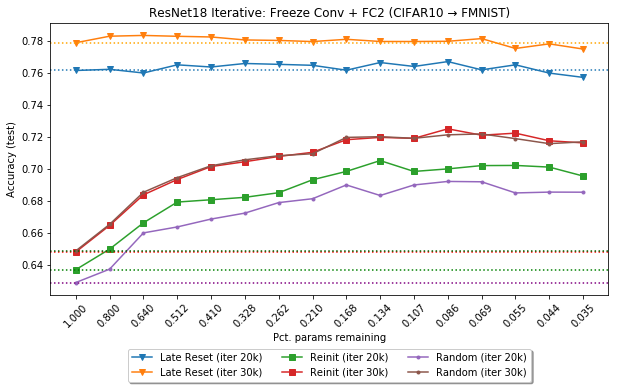}
    \caption[Winning Tickets: CIFAR-10 to FMNIST (Freeze + FC2)]{Winning Tickets: CIFAR-10 to FMNIST (ResNet18, freeze conv. + 2-layer fully-connected)}
    \label{fig:ticket_freeze_fc2_fmnist}
\end{figure}

We repeat our previous experiment, this time with FashionMNIST as the target dataset. Again, we find winning tickets during training on the source task, and fine-tune the fully-connected layers with the pruned convolution layers frozen. As before, we replace the single output layer in ResNet18 with a 2-layer fully-connected network. Figure \ref{fig:ticket_freeze_fc2_fmnist} summarizes our results.

Again, we observe the same phenomenon as before; the sub-networks that are reset to $m \odot \bm{\theta}_S$ outperform all other initializations. We find winning tickets with as few as 7\% of original parameters remaining. In addition, the previous effect we noticed with respect to reinitialization holds; all sub-networks reset to $m \odot \bm{\theta}_0$ or $m \odot \bm{\theta}'$ fail to reach commensurate accuracy. However, as the sparsity in the convolution layers increases, test accuracy improves.

\subsection{Reinitialization dynamics with frozen convolutions}
A puzzling feature of both experiments above is that although sparse sub-networks never reach commensurate accuracy when trained from the winning ticket initialization $m \odot \bm{\theta}_0$, the accuracy improves and reaches a plateau as sparsity increases when trained from both the winning ticket as well as random reinitialization. 

Since the convolution layers are frozen during fine-tuning on the target task, re-initializing these layers either to $m \odot \bm{\theta}_0,$ or $m \odot \bm{\theta}'$ for any $\bm{\theta'} \in \mathcal{D}_\theta$ should adversely affect accuracy. This is true, but accuracy also increases as we prune the convolution filters, without any noticeable drop-off. 

We hypothesize that as we prune the layers following a reinitialization, we simply approach the performance of the fully-connected layers trained in isolation on the dataset. We experimentally test this by measuring the performance of winning ticket and random initializations relative to two networks; a dense, 2-layer fully-connected network, and a (single-layer) logistic regression classifier.

\begin{figure}[h]
    \centering
    \includegraphics[width=0.75\textwidth]{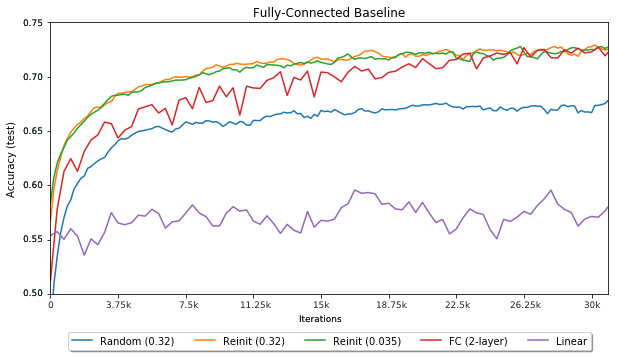}
    \caption[Training Dynamics: Comparison to Fully-Connected Baseline]{Comparison of CIFAR-10 transferred to FMNIST with 1 and 2-layer networks trained from scratch.}
    \label{fig:baseline_fmnist}
\end{figure}

Figure \ref{fig:baseline_fmnist} shows our results for this experiment. In particular, we note that the 2-layer network approaches the performance of ResNet18 when the convolution layers are frozen and reset during fine-tuning, while the linear regression baseline hovers between 58-65\% accuracy. This confirms our hypothesis; as we increase the sparsity of the model, retraining our pruned network from its original initialization matches the performance of a 2-layer fully-connected network trained from scratch. This suggests that when reinitialized, any winning tickets found in the original deep network will effectively match the performance of training the classification layers in isolation. We note this also holds when we randomly reinitialize our winning ticket.

In the next section, we will show that if we eliminate the restriction that we can only fine-tune the convolution layers, we can recover performant networks trained from the winning ticket initialization.

\section{Fine-Tuning the Entire Network}\label{sec:vision_lt_ft_all}
In this section, we relax the previous constraint and also train the convolution layers when fine-tuning on the target task. In particular, we examine the performance of iterative vs. one-shot pruning for ResNet18, and find winning tickets with as few as 7\% of parameters remaining for both NORB and FMNIST. We also show that iterative pruning {\it fails} to produce winning tickets at {\it any} pruning level for VGG19 when transferring CIFAR-10 to both NORB and FMNIST; for brevity's sake, we defer the full results to the appendix.

\subsection{Transferring Winning Tickets for ResNet18}
\begin{figure}[h]
    \centering
    \singlespacing
    \includegraphics[width=0.75\textwidth]{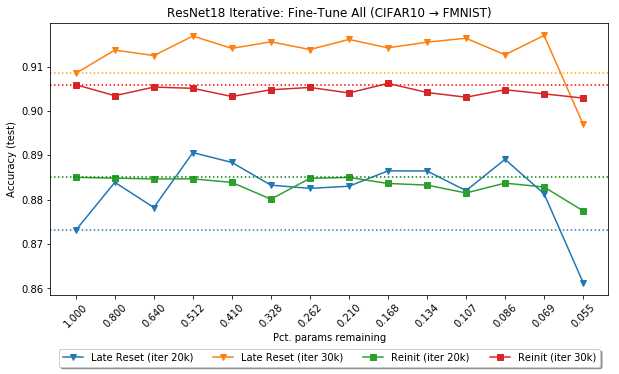}
    \caption[Winning Tickets: CIFAR-10 to FMNIST (ResNet18, fine-tune all)]{Winning Tickets: CIFAR-10 to FMNIST (ResNet18, fine-tune all, iterative pruning)}
    \label{fig:resnet_iterative_all_fmnist}
\end{figure}

We begin with our results for ResNet18. Figure \ref{fig:resnet_iterative_all_fmnist} shows our results for transferring winning tickets found for CIFAR-10 to FMNIST. 

When we are able to fine-tune the convolution layers alongside the fully-connected layers, we find winning tickets with 6.9\% of parameters remaining for ResNet18 when our network is initialized to $m \odot \bm{\theta}_S$. After 30k and 50k training steps respectively, the late-reset sub-network exceeds original network's accuracy at all pruning levels until 6.9\%. While the winning tickets we find are the same size as those when we freeze the convolution layers during fine-tuning, we see that fine-tuning the convolutional layers improves the test accuracy of our best winning ticket to 91.7\%.

Moreover, we see again that re-initializing the network to its original weights $m \odot \bm{\theta}_0$ adversely affects the network's performance on down-stream tasks. After 50k iterations, we find that winning tickets reset to $m \odot \bm{\theta}_0$ fail to reach commensurate test accuracy at any sparsity level. However, the performance of the re-initialized winning ticket massively improves when we are allowed to fine-tune the convolutional layers. 

\subsection{Iterative vs. One-Shot Pruning}
\cite{Frankle18} show that iterative pruning consistently finds the smallest and most accurate winning tickets for a wide variety of simple models. Accordingly, we investigate the performance of one-shot vs. iterative pruning for finding winning tickets when transferring CIFAR-10 to SmallNORB.

\begin{figure}[h]
    \centering
    \singlespacing
    \includegraphics[width=0.75\textwidth]{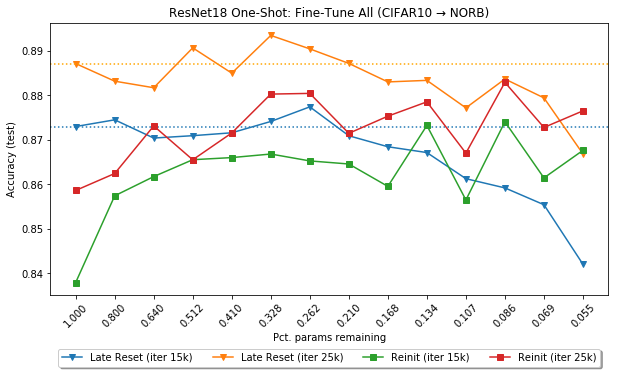}
    \caption[Winning Tickets: CIFAR-10 to NORB (ResNet18, one-shot)]{Winning Tickets: CIFAR-10 to NORB (ResNet18, fine-tune all, one-shot pruning)}
    \label{fig:resnet_oneshot_all_norb}
\end{figure}

Figure \ref{fig:resnet_oneshot_all_norb} shows our results for transferring winning tickets found for CIFAR-10 with one-shot pruning to NORB. Sub-networks initialized at $m \odot \bm{\theta}_S$ match commensurate test accuracy to an extent; the smallest winning ticket we find is with 21.0\% of parameters remaining, and the most accurate is found with 32.8\% remaining (with a test accuracy of 89.3\%). Moreover, when the network is reset to $m \odot \bm{\theta}_0$, we fail to find winning tickets at all sparsity levels.

\begin{figure}[h]
    \centering
    %TODO: FIX
    \includegraphics[width=0.75\textwidth]{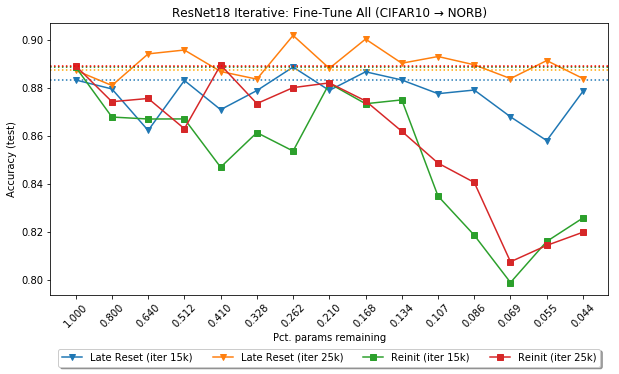}
    \caption[Winning Tickets: CIFAR-10 to NORB (ResNet18, iterative)]{Winning Tickets: CIFAR-10 to NORB (ResNet18, fine-tune all, iterative pruning)}
    \label{fig:resnet_iterative_all_norb}
\end{figure}

In comparison, Figure \ref{fig:resnet_iterative_all_norb} shows the results for ResNet18, when winning tickets are found with iterative pruning. The sub-networks reset to $m \odot \bm{\theta}_S$ again outperform all other initializations, and exceed the commensurate accuracy at nearly all pruning levels. The smallest winning ticket we find has 5.5\% of parameters remaining, while the winning ticket that reaches the top validation accuracy has 26.2\% of parameters remaining, and reaches a test accuracy of 90.2\%. As before, all networks reset to $m \odot \bm{\theta}_0$ fail to reach commensurate accuracy. Therefore, iterative pruning boosts both sparsity as well as accuracy.

\section{Conclusion}

In summary, we extend the Lottery Ticket Hypothesis to a number of transfer learning tasks in vision. We pose the {\it Ticket Transfer Hypothesis,} and experimentally validate it by showing that sparse sub-networks found with unstructured pruning can be transferred to similar tasks in various settings. Specifically, we find winning tickets both (1) when we fine-tune the fully-connected layers while freezing the convolution layers, and (2) when we fine-tune the entire network end-to-end. Moreover, we confirm the efficacy of iterative pruning, and show that winning tickets from VGG19 do not transfer well to down-stream tasks.

\section*{Acknowledgements}
I am grateful to Sanjeev Arora for his advice and guidance over the course of this work. I would also like to thank Jonathan Frankle, Pravesh Kothari, Karthik Narasimhan, Anthony Bak, and Matt Elkherj for many fruitful conversations throughout the preparation of this manuscript. This work was made possible by a generous grant from Amazon Research for our compute infrastructure.

\bibliographystyle{plainnat}
\bibliography{main}

% ==========START APPENDICES==============
\appendix

\section{Transferring Winning Tickets for VGG19}
We repeat a subset of our experiments for VGG19. We conduct the lottery ticket experiment for both NORB and FMNIST, and examine both the late-reset ($m \odot \bm{\theta}_S$) and winning ticket ($m \odot \bm{\theta}_0$) initializations. 

\begin{figure}[h]
    \centering
    \includegraphics[width=0.75\textwidth]{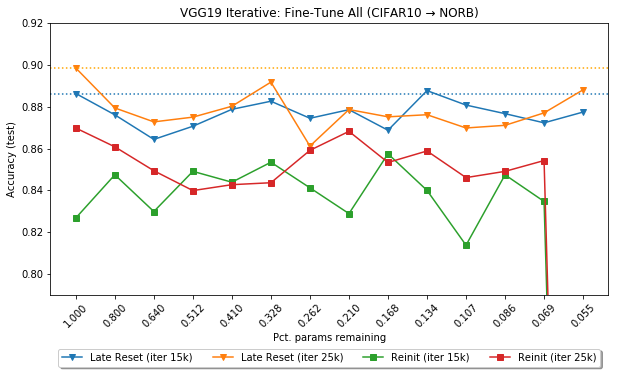}
    \caption[Winning Tickets: CIFAR-10 to NORB (VGG19, fine-tune all)]{Winning Tickets: CIFAR-10 to NORB (VGG19, fine-tune all, iterative pruning)}
    \label{fig:vgg_iterative_all_norb}
\end{figure}

Figure \ref{fig:vgg_iterative_all_norb} shows the results of the lottery ticket experiment for VGG19 when transferring CIFAR-10 to NORB. At all sparsity levels, the sparse sub-networks fail to reach commensurate test accuracy. Networks reset to the winning ticket initialization still perform worse than the late-reset initialization, but accuracy notably degrades as sparsity increases.

\begin{figure}[h]
    \centering
    \includegraphics[width=0.75\textwidth]{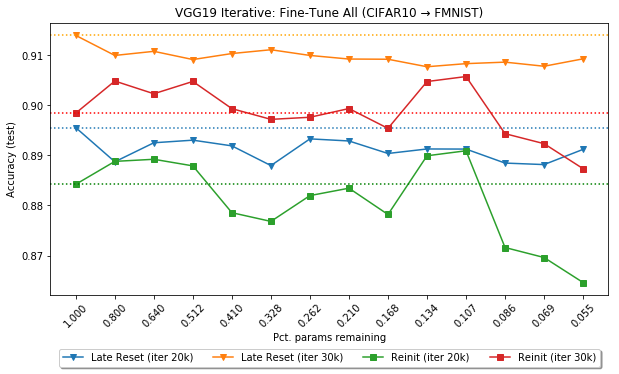}
    \caption[Winning Tickets: CIFAR-10 to FMNIST (VGG19, fine-tune all)]{Winning Tickets: CIFAR-10 to FMNIST (VGG19, fine-tune all, iterative pruning)}
    \label{fig:vgg_iterative_all_fmnist}
\end{figure}

We observe similar results when we repeat this experiment on FMNIST in Figure \ref{fig:vgg_iterative_all_fmnist}. We similarly observe that we fail to find winning tickets at all sparsity levels; for both the late-reset and winning ticket initializations, accuracy decreases as we increase the sparsity.

This corroborates the findings of \cite{Frankle18} and \cite{frankle19scale}; they show that variants of ResNet consistently produce smaller and more accurate winning tickets than VGG, and only focus on ResNet-type architectures when scaling winning tickets up to ImageNet. These results provide another piece of evidence that ResNet produces superior winning tickets. In Section \ref{sec:future_work}, we draw on the connection between ResNets and ensemble learning identified by \cite{VeitWB16} to provide some intuition as to why this might be the case.

\section{Related Work}

We review several lines of research related to our problem. We first summarize various approaches for sparsifying neural networks, including regularization and pruning. We formally state the Lottery Ticket Hypothesis of \cite{Frankle18}, and finally provide a high-level overview of transfer learning.

\subsection{Network Sparsity}
Since initial neural network architectures were proposed, the closely-related problem of reducing the parameter count while maintaining accuracy has been extensively investigated. Benefits include lowering the memory footprint of the model, speeding up prediction, and potentially-improved generalization performance. Sparsification in this regime is also of independent interest, as it can shed light on the underlying dynamics of more complex, overparameterized models. In recent years,  three dominant approaches have emerged:

\paragraph{Regularization}
The concept of regularization has a long history which extends far outside the domain of training deep neural networks. Various forms of regularization were simultaneously invented for solving ill-posed optimization problems and preventing overfitting. One of the earliest forms is known as Tikhonov regularization, which imposes an $\ell_2$ weight penalty on model parameters. In classical statistics, this is called {\it ridge regression}, and is connected to {\it weight-decay} in the machine learning community. Other forms of ``regularized regression" include an $\ell_1$ penalty, commonly referred to as LASSO. For an in-depth summary of regularization in classical statistics, see \cite{hastie15}.

This concept translates naturally to deep neural networks, with a small modification; instead of penalizing the $\ell_2$ norm of all weights in the loss function, {\it decay} the gradient update at each iteration proportional to the $\ell_2$ norm of the particular parameter. \cite{alexnet} cite a small weight-decay parameter as a crucial component of AlexNet's state-of-the-art performance at ILSVRC `12 . The same procedure can be performed with the $\ell_1$ norm instead, which imposes sparsity on the fitted weights \citep{kukacka}.

More recent types of regularization directly seek to impose sparsity during training. The most well-known family of these methods is called {\it dropout regularization}, which was introduced by \cite{srivastava14a}. This method stochastically ``drops out" certain inputs in each layer throughout training, which prevents overfitting and improves test accuracy. Recent extensions of this technique include variational dropout \citep{kingma15dropout, molchanov17}. \cite{louizos17} propose $\ell_0$ regularization, a related approach which derives a novel technique to circumvent the non-differentiability of the $\ell_0$ norm and imposes a strict sparsity constraint during training.

\paragraph{Task-Specific Architectures}
Another approach to reducing the parameter count of models and improving inference speed is developing streamlined, task-specific architectures. Rather than seeking to sparsify an existing deep network, these models are engineered specifically with parameter savings and inference speed in mind. 

Two examples in this category that still remain state-of-the-art for many use-cases include SqueezeNet \citep{SqueezeNet} and MobileNetV1/V2 \citep{mobilenet, mobilenetv2}. SqueezeNet incorporates techniques such as replacing $3 \times 3$ filters with $1 \times 1$ filters and delaying pooling until the final stages of the network to increase activation magnitude in the convolution layers. MobileNetV1 leverages the insight that a convolution can be expressed as a composition of {\it depth-wise separable} and {\it point-wise} convolutions, and introduces two hyper-parameters to control the {\it width} of the network and the {\it resolution} of the input images \cite{mobilenet}. MobileNetV2 extends its predecessor by incorporating residual blocks (such as in \cite{HeZRS15}) along with depth-wise separable convolutions, and achieves state-of-the-art performance for a number of inference tasks while approaching the accuracy of very deep, dense networks such as ResNet and VGG \cite{mobilenetv2}.

\paragraph{Network Pruning}
A third approach to network sparsity (and the one that we will predominantly focus on in this thesis) is known as {\it network pruning}. In this setting, the full model is first trained end-to-end, and subsequently parameters are removed via a heuristic. The model is then re-trained after pruning to recover from the loss in accuracy. 

Pruning and related concepts were introduced as early as 1990 by \cite{Cun90optimalbrain} through the concept of optimal brain damage. \cite{HanPTD15} and \cite{LiKDSG16} provide two modern approaches to network pruning. \cite{HanPTD15} introduce a methodology to prune networks with one round of fine-tuning (i.e. train, prune, then fine-tune) that achieved significant parameter savings on VGG (about a 50\% parameter reduction). They also extend this to {\it iterative} pruning, which performs multiple rounds of fine-tuning. Broadly, these methods are referred to as {\it unstructured} pruning, since all layers are pruned simultaneously by the heuristic.

Similarly, \cite{LiKDSG16} introduce {\it structured} pruning, in which entire layers/structured subsets of layers are pruned prior to fine-tuning; they describe a method for pruning CNNs in which whole filters are pruned. The filter-based pruning reduces the parameter count by about 38\% for ResNet-110 without an appreciable drop in accuracy. It is important to note here that while unstructured, magnitude-based pruning can find much smaller networks that reach the same accuracy, structured pruning yields computational speedups as well by making the network shallower.

\cite{Frankle18} use the two unstructured pruning methods introduced by \cite{HanPTD15} In particular, they use both {\it iterative} and {\it one-shot} pruning to find sparse, trainable sub-networks. All results are found by using simple magnitude-based pruning, in which the smallest $p\%$ of weights by magnitude are pruned at each iteration. For a summary of current state-of-the-art approaches to pruning, see \cite{gale2019state}. Pruning has also been investigated in the context of transfer learning by \cite{molchanov17}, who describe a method for pruning entire filters based on the first-order gradient information of the layers to improve the generalization performance of ResNet-50 for transfer learning tasks.

\paragraph{Pruning as Network Architecture Search} Another interpretation of network pruning frames it as a form of {\it network architecture search}. Conventional wisdom holds that widely-used network architectures are massively over-parameterized; this overparameterization aids in generalization performance, and has been shown to also positively affect optimization in certain circumstances \citep{AroraOprm18}. Accordingly, several authors take the view that pruning uncovers hidden networks within a massively over-parameterized deep network. \cite{morphnet} propose MorphNets, which learn a sparse architecture throughout the training process via a series of interleaved {\it prune} and {\it expand} phases. Similarly, \cite{He18AMC} develop an automatic approach to finding sparse architectures which they call ``AutoML for Model Compression" (AMC). They frame the problem of coarse, structured pruning as a reinforcement learning problem and describe an automatic pipeline for pruning channels and fine-tuning. 

\subsection{The Lottery Ticket Hypothesis}

\cite{Frankle18} introduced the Lottery Ticket Hypothesis, and provided a wide set of experimental evidence including experiments on basic fully-connected networks as well as convolutional networks trained on CIFAR-10. They also posed the question of whether the sparse representation produced by pruning can be transferred between tasks. Subsequently, \cite{frankle19scale} showed that a slight modification of the above method finds winning lottery tickets for ResNet50 trained on ImageNet. The authors achieve test accuracy within a percentage point of the un-pruned model with as many as 79\% of original parameters removed. In particular, they introduce {\it late resetting,} in which the sparse sub-network is re-initialized not to $m \odot \bm{\theta}_0$, but rather $m \odot \bm{\theta}_t$, for some $t \ll j$ (recall that $j$ is the iteration at which the original network $f(\cdot)$ reaches its minimum validation accuracy). By resetting the network to a checkpoint several epochs {\it after} initialization, they find strong evidence of winning tickets for large-scale visual recognition problems. In a recent follow-up work, \cite{deconstructingLT} investigate some of the underlying dynamics of winning tickets, and find that masking can be viewed as a form of training, insofar that it biases the retained weights in a particular direction. \cite{lee2018snip} introduce a distinct but related approach called SNIP, which performs one-shot pruning on the network based on the sensitivity of individual connections. However, while exceeding the performance of randomly-reinitialized lottery tickets, the ``winning ticket initialization" discovered by \cite{Frankle18} outperforms SNIP by an appreciable margin.\footnote{See \href{https://openreview.net/forum?id=rJl-b3RcF7}{\tt https://openreview.net/forum?id=rJl-b3RcF7} for details.}

\subsection{Evidence against the Lottery Ticket Hypothesis}
While there is a sufficient body of evidence in favor of the Lottery Ticket Hypothesis, other works have focused on a holistic comparison of various pruning heuristics and methods. In particular, \cite{gale2019state} conduct an extensive investigation into various methods for inducing sparsity, including variational dropout due to \cite{molchanov17}, $\ell_0$ regularization due to \cite{louizos17}, and unstructured magnitude pruning as originally described by \cite{HanPTD15}. While they demonstrate that naive magnitude pruning outperforms more complex methods, they also find that naive magnitude pruning (without late-resetting) fails to find winning tickets for both ResNet50 on ImageNet, as well as for Transformer-based models on the WMT 2014 English-to-German translation task.

In addition, \cite{liu18} consider network pruning from the perspective of network architecture search and show that, among other results, a careful choice of learning rate can alleviate the negative effects of random re-initialization. In particular, they show that with a carefully-chosen learning rate, the ``winning ticket initialization" provides no appreciable benefit over random initialization for the pruned sub-network. 

\section{Limitations \& Future Work}\label{sec:future_work}

Our results in Sections \ref{sec:vision_lt_freeze} and \ref{sec:vision_lt_ft_all} rely on small-scale datasets such as CIFAR-10, SmallNORB, and FashionMNIST. In a similar manner to \cite{frankle19scale}, we would like to scale up our existing experiments to deeper models such as ResNet-50/ResNet-152 trained on ImageNet. We also believe that larger-scale tasks will show a far greater accuracy drop-off when using the winning ticket or random re-initialization, since fully-connected networks cannot begin to approach the accuracy of convolutional networks on larger-scale tasks. In addition, we also believe that the Ticket Transfer Hypothesis offers a number of interesting directions for future research in network sparsity.

\paragraph{Transferring winning tickets between tasks} In all our experiments, we considered a transfer learning setting in which we fine-tune the pre-trained network on a different dataset but on the same task (object recognition). It would be interesting to verify if winning tickets found for a source task such as object detection could transfer to distinct down-stream tasks such as semantic segmentation or image captioning.
 
\paragraph{Winning tickets via structured pruning} \cite{Frankle18} posed the original Lottery Ticket Hypothesis in the context of unstructured magnitude pruning. While we are able to obtain massive parameter savings, pruned models cannot take advantage of this increased sparsity to improve prediction speed, unless specialized hardware is used \citep{HanPTD15}. Another approach to pruning is given by \cite{LiKDSG16}, which proposed {\it structured layer-wise} pruning which eliminates whole convolutional filters. It is interesting to ask if pruning ResNet and VGG in this manner will produce winning tickets that are more computationally efficient for transfer learning.

\paragraph{Connection to ensemble learning} In addition to scaling up our experiments to larger tasks in different contexts, we also propose an avenue towards determining the underlying dynamics of the lottery ticket hypothesis. In particular, both \cite{Frankle18} and \cite{frankle19scale} determine that winning tickets found for ResNet-type architectures outperform VGG for vision tasks. Moreover, our results in Section \ref{sec:vision_lt_ft_all} corroborate this finding, as winning tickets found on VGG fail to reach commensurate accuracy at all sparsity levels. We hypothesize that the improved performance of ResNet in these tasks can be accounted for by its implicit training dynamics as an ensemble; in particular, \cite{VeitWB16} show that ResNet behaves like an ensemble of shallow residual networks during training. This view of ResNet as an {\it overparameterized ensemble} (with weight-sharing) leads to a natural interpretation of our unstructured pruning; we prune all the weak learners in the overparameterized ensemble simultaneously. This also might suggest why winning tickets found on ResNet transfer well to other tasks. To this end, we ask if a lottery ticket-style experiment can be devised on a traditional ensemble model (such as a random forest) to investigate this connection.

\clearpage
\section{Network Configurations}

\begin{table}[h]
\small
\begin{center}
\begin{tabular}{|l|cc|c|c|}
\hline
\multicolumn{1}{|l|}{\it Network} & \multicolumn{2}{c|}{Fully-Connected} & ResNet18& VGG19 \\ \hline
\textit{Convolutions} & \multicolumn{2}{c|}{}& \begin{tabular}[c]{@{}c@{}}16 3x{[}16,16{]}\\ 3x{[}32,32{]}\\ 3x{[}64,64{]}\end{tabular}& \begin{tabular}[c]{@{}c@{}}2x64, pool 2x128\\ pool, 4x256, pool\\ 4x512, pool, 4x512\end{tabular} \\ \hline
\textit{FC Layers}& 1024, 10 & 1024, 100, 10 & avg-pool, 10& avg-pool, 10\\ \hline
\textit{All/Conv Weights} & 10,240/0 & 1.02M/0 & 11.6M/10.9M & 20.03M/20.02M \\ \hline
\textit{Iterations/Batch} & 30k/128& 30k/128 & \begin{tabular}[c]{@{}c@{}}30k/128 CIFAR-10\\ 25k/128 NORB\\ 30k/128 FMNIST\end{tabular} & \begin{tabular}[c]{@{}c@{}}112k/64 CIFAR-10\\ 50k/64 NORB\\ 112k/64 FMNIST\end{tabular}\\ \hline
\textit{Optimizer}& \multicolumn{2}{c|}{\begin{tabular}[c]{@{}c@{}}SGD 0.01,0.005, \\ Momentum 0.9\end{tabular}} & \begin{tabular}[c]{@{}c@{}}SGD 5e-3,1e-3,1e-4,\\ Momentum 0.9,\\ Weight Decay 1e-4\end{tabular} & \begin{tabular}[c]{@{}c@{}}SGD 1e-2-1e-3,1e-4,\\ Momentum 0.9,\\ Weight Decay 1e-4\end{tabular} \\ \hline
\textit{Pruning Rate} & \multicolumn{2}{c|}{N/A} & \multicolumn{2}{c|}{Conv: 20\%, FC: 0\%}\\ \hline
\end{tabular}
\vspace{2mm}
\caption{Model configurations for vision experiments}\label{tab:vision_models}
\end{center}
\end{table}

\setlength{\tabcolsep}{6pt}
\renewcommand{\arraystretch}{1.5}
\begin{table}[h]
\small
\begin{center}
\begin{tabular}{|lc|cc|}
\hline
\textit{Domain}       & \textbf{Source} & \multicolumn{2}{c|}{\textbf{Target}} \\ \hline
\textit{Dataset}      & CIFAR-10        & SmallNORB             & FMNIST            \\ \hline
\textit{Train/Test}   & 50k/10k         & 40k/10k          & 60k/10k           \\ \hline
\textit{Num. class} & 10              & 5                & 10                \\ \hline
\textit{Shape}    & 32x32x3   & 28x28x1    & 28x28x1    \\ \hline
\textit{Task}         & \multicolumn{3}{c|}{Multi-Class Classification}        \\ \hline
\end{tabular}
\vspace{2mm}
\caption[Summary of vision datasets/tasks]{Summary of source and target datasets and tasks for vision experiments.}\label{tab:vision_datasets}
\end{center}
\end{table}
% Reset row/colsep for subsequent tables
\setlength{\tabcolsep}{6pt}
\renewcommand{\arraystretch}{1}

\end{document}